\documentclass[10pt, a4paper]{article}

\usepackage{lrec-coling2024} 

\usepackage{xstring}
\usepackage{color}
\usepackage{booktabs}
\usepackage{xcolor}
\usepackage{hyperref}
\usepackage{fontawesome5}
\usepackage{multirow}

\title{How to Encode Domain Information in Relation Classification}

\name{
Elisa Bassignana\textsuperscript{\faCompass}\textsuperscript{\faRobot} \hspace{1em}
Viggo Unmack Gascou\textsuperscript{\faCompass} \hspace{1em}
Frida Nøhr Laustsen\textsuperscript{\faCompass} \hspace{1em} \\
{\bf \large 
Gustav Kristensen\textsuperscript{\faCompass} \hspace{1em}
Marie Haahr Petersen\textsuperscript{\faCompass} \hspace{1em}
Rob van der Goot\textsuperscript{\faCompass}\textsuperscript{\faRobot}} \\
{\bf \large 
Barbara Plank\textsuperscript{\faCompass}\textsuperscript{\faMountain}}} 

\address{
\textsuperscript{\faCompass}IT University of Copenhagen \hspace{1em}
\textsuperscript{\faRobot}Pioneer Centre for AI \hspace{1em}
\textsuperscript{\faMountain}LMU Munich \\
\texttt{\{elba, viga, fril, gukr, mhpe, robv\}@itu.dk} \hspace{1em}\texttt{b.plank@lmu.de}\\
}

\abstract{
Current language models require a lot of training data to obtain high performance.
For Relation Classification (RC), many datasets are domain-specific, so combining datasets to obtain better performance is non-trivial.
We explore a multi-domain training setup for RC, and attempt to improve performance by encoding domain information.
Our proposed models improve $> 2$ Macro-F1 against the baseline setup, and our analysis reveals that not all the labels benefit the same: The classes which occupy a similar space across domains (i.e., their interpretation is close across them, for example \textit{physical}) benefit the least, while domain-dependent relations (e.g., \textit{part-of}) improve the most when encoding domain information.
 \\ \newline  }

\begin{document}

\maketitleabstract

\section{Introduction}

Relation Classification (RC) is the task of identifying the semantic relation between two given entities. The task is beneficial for many different downstream tasks which involve Natural Language Understanding. For example, question answering, knowledge base population, or summarization.
In addition to the wide variety of downstream applications, as most information extraction tasks, RC is topic-specific, meaning that depending on the topic the information to extract can vary a lot. For example, in the music domain we may want to extract that a song is included in a musical album, while in the politics domain we may have a politician winning a political election.
While current deep learning models require a lot of training data, collecting and annotating text from every domain is time-consuming and expensive.

In this project, we explore the critical setup of multi-domain training with the aim of identifying the best setup for maximizing the training data (by including data coming from different domains), without losing domain-specific information.
To do so, we compare multiple ways of enriching the input instances with domain information (see Section~\ref{sec:domains}).

Encoding information about where a certain utterance originates from has been previously explored in other Natural Language Processing fields.
In the multi-lingual space,~\citet{xlmr} exploited language embeddings for multi-lingual model training.
\citet{ammar-etal-2016-many} first proposed to use language embeddings for training a multi-lingual syntactic parser for seven European languages, and showed improved performance.
Later work also successfully trained parsers with the so called treebank embeddings for datasets within the same language~\cite{stymne-etal-2018-parser} or language family~\cite{smith-etal-2018-82}. 
Other work have used special language ids to mark the language of each instance in the context of machine translation~\cite{liu-etal-2020-multilingual-denoising}.
To the best of our knowledge, these approaches have been exploited mostly in multi-lingual setups and syntactic tasks. In this work, we explore a gap and test their effectiveness for encoding domain information in a semantic setup: Relation Classification. 
We compare ``dataset embeddings'' and ``domain markers'' from previous work with a new approach exploiting domain-specific entity types.
Our contributions are:

\begin{itemize}
    \item CrossRE 2.0, an extension of the CrossRE dataset~\cite{bassignana-plank-2022-crossre} with 3.3k new annotations in the news domain in order to balance data across domains;
    \item We propose the first multi-domain training baseline on CrossRE;
    \item We test previous work for encoding dataset information in RC, and compare it with a new RC-specific technique; We present an in-depth analysis of the results obtained.
\end{itemize}

\section{Domain Encoding for Relation Classification}
\label{sec:domains}

\subsection{Dataset Embeddings}

The dataset embedding model tries to encode information about the domain with ad-hoc embeddings on the encoder side.
Dataset embeddings are vector representations learned at training time that aim at capturing distinctive properties of multiple data sources into a continuous vector, without losing their heterogeneous characteristics. 
Originally they were often concatenated to the word embedding and then used in e.g., a Bi-LSTM~\cite{stymne-2020-cross,wagner-etal-2020-treebank,van-der-goot-etal-2021-effectiveness}. However, since large language models have become the standard, this has become trickier, as they have a pre-determined input size. To enable usage of dataset embeddings,~\citet{van-der-goot-de-lhoneux-2021-parsing} propose to sum them to the input representation. 
In our setup, we treat each domain as a separate data source.

\subsection{Special Domain Markers}
An intuitive and simple alternative way of encoding the domain is by using special tokens appended to the input text itself. 
This has been previously done in  machine translation in order to mark the different languages~\cite{liu-etal-2020-multilingual-denoising}.
We concatenate a special token at the beginning of each instance containing the corresponding domain (e.g.,~\texttt{[MUSIC]} or \texttt{[NEWS]}).
These domain markers are treated by the tokenizer as special tokens, i.e., they are not tokenized into subwords, so the model learns a representation for each of them during training.

\subsection{Entity Type Information}

The domain-specific entity types carry out information which can be relevant for identifying the correct relation label.
Following~\citet{zhong-chen-2021-frustratingly} we add entity type information in the representation of the input (see model description in Section~\ref{sec:model}).
We test two different approaches to do this.
First, we use the 39 fine-grained types proposed by~\citet{crossNER} including e.g., \textit{musician} or \textit{political party}, which are domain-specific.
In the second setup we map these fine-grained types into five coarse-grained types. For example, \textit{musician} and \textit{political party} are mapped to \textit{person} and \textit{organization} respectively. While this last approach shades domain information, it guarantees a more condensed entity type distribution, and it can be combined with the other two setups.

\section{Experimental Setup}

\subsection{Data}
\label{sec:data}

CrossRE~\cite{bassignana-plank-2022-crossre},\footnote{Released with a GNU General Public License v3.0.} is a manually-annotated dataset for multi-domain RC including 17 relation types spanning over six diverse text domains: news (\faNewspaper), politics (\faLandmark), natural science (\faLeaf), music (\faMusic), literature (\faBookOpen), and artificial intelligence (\faRobot). The dataset was annotated on top of CrossNER~\cite{crossNER}, a Named Entity Recognition (NER) dataset.
Table~\ref{tab:dataset-statistics} reports the statistics of CrossRE.
While the train, dev, and test splits include similar amounts of sentences across the six domains, the number of annotated relations varies over a wider range. The reason for this includes different average sentence lengths, and different relation densities across the domains. In the original dataset, the news domain is particularly small. This domain comes from a different source with respect to the other five---the CoNLL-2003 dataset~\cite{tjong-kim-sang-de-meulder-2003-introduction} and Wikipedia~\cite{crossNER} respectively.

\begin{table}
    \centering
    \resizebox{\columnwidth}{!}{
    \begin{tabular}{r|rrr|r|rrr|r}
    \toprule
    & \multicolumn{4}{c|}{\textsc{sentences}} & \multicolumn{4}{c}{\textsc{relations}} \\
    \midrule
    & train & dev & test & \textbf{tot.} & train & dev & test & \textbf{tot.} \\
    \midrule
    \faNewspaper & 217 & 1,320 & 3,053 & 4,590 & 156 & 1,043 & 2,115 & 3,314 \\
    \midrule
    \faNewspaper & 164 & 350 & 400 & 914 & 175 & 300 & 396 & 871 \\
    \faLandmark & 101 & 350 & 400 & 851 & 502 & 1,616 & 1,831 & 3,949 \\
    \faLeaf & 103 & 351 & 400 & 854 & 355 & 1,340 & 1,393 & 3,088 \\
    \faMusic & 100 & 350 & 399 & 849 & 496 & 1,861 & 2,333 & 4,690 \\
    \faBookOpen & 100 & 400 & 416 & 916 & 397 & 1,539 & 1,591 & 3,527 \\
    \faRobot & 100 & 350 & 431 & 881 & 350 & 1,006 & 1,127 & 2,483 \\
    \midrule
    \textbf{tot.} & 885 & 3,471 & 5,499 & \textbf{9,855} & 2,431 & 8,705 & 10,786 & \textbf{21,922} \\
    \bottomrule
    \end{tabular}}
    \caption{\textbf{CrossRE 2.0 Statistics.} Number of sentences and number of relations of the news extension (first row), and statistics of the original domains of CrossRE (below).
    }
    \label{tab:dataset-statistics}
\end{table}

\subsubsection{CrossRE 2.0}

With the aim of mitigating the effect of dataset size on the model performance, which influences the comparison of results across domains, we expand the news domain of CrossRE.
We follow the annotation guidelines by~\citet{bassignana-plank-2022-crossre}\footnote{\url{https://github.com/mainlp/CrossRE/blob/main/crossre_annotation/CrossRE-annotation-guidelines.pdf}} and manually annotate more than 4.5k sentences from the CoNLL-2003 dataset~\cite{tjong-kim-sang-de-meulder-2003-introduction}---the original data source of this domain. The data is annotated by a hired linguist compensated fairly according to national salary scales, who got extensive training for the task.
We refer to~\citet{bassignana-plank-2022-crossre} for the discussion on the annotation agreement because for consistency we employed the same annotator who annotated the original version of CrossRE.
Table~\ref{tab:dataset-statistics} reports the statistics of our extension, with over 3k annotated relations and an overall total in news (including the original dataset) of 4.1k, which is in line with the other domains. The dataset extension is publicly available in the CrossRE repository.\footnote{\url{https://github.com/mainlp/CrossRE/}} We train the model in a multi-domain setup, i.e., mixing the six training sets of CrossRE 2.0.

\subsection{Model Architecture}
\label{sec:model}

We use the baseline model of the original CrossRE paper.\footnote{\url{https://github.com/mainlp/CrossRE}} Following the model architecture first proposed by \citet{baldini-soares-etal-2019-matching}, the implementation by~\citet{bassignana-plank-2022-crossre} augments the sentence with four entity markers $e_1^{start}$, $e_1^{end}$, $e_2^{start}$, $e_2^{end}$ surrounding the two entities.
When exploiting the entity types, the information is injected in the entity markers (e.g., \texttt{[E1:person]})
The augmented sentence is then passed through a pre-trained encoder, and the classification is made by a linear layer over the concatenation of the start markers $[\hat{s}_{e_1^{start}}, \hat{s}_{e_2^{start}}]$.
We run our experiments over five random seeds and report the average. 
All hypermarameters follow~\citet{bassignana-plank-2022-crossre}.
Our code is available on GitHub.\footnote{Project repository \url{https://github.com/viggo-gascou/multi-domain-rc}}

\section{Results}
\label{sec:results}

Table~\ref{tab:results} reports the Macro-F1 results of our experiments.
The dataset embeddings setup fails to beat the baseline. 
The main reason for this is the limited amount of training data in our setup, which challenges the model in learning them. \footnote{We manually inspected the dataset embeddings before and after training.} 
The dataset embeddings are summed to the word, segment, and position embeddings, which are then updated all at once in the forward pass.
Additionally, in settings where they are successful, these embeddings are usually used to disambiguate datasets coming from difference data sources or languages. Here instead we are at a more fine-grained level, trying to model different topics, with data extracted from the same source (except for news).

Concatenating a special domain marker at the beginning of the sentence results in the best performance (36.90 Macro-F1), with the highest improvement in the music domain (+2.96) and sometimes small yet consistent improvements across all domains.
The fine-grained entity types lead to decreased performance, because their distribution is very sparse across the six domains. For example, out of the 39, the news domain from CoNLL 2003 only includes \textit{person}, \textit{location}, \textit{organization} and \textit{miscellaneous}, resulting in a performance decrease of -1,78.
Using the coarse-grained entity types---shared across all the domains---results in a slightly better average Macro-F1 (34.99) than employing the fine-grained ones (34.10), but it does not improve over the baseline either.
As this setup lacks domain information, we try combining the coarse-grained entity representation with the special domain markers. Within this setup results are mixed across the domains: While most of them (except AI) improve over the coarse-grained entity type (without domain information), only politics, science, music and literature overcome the baseline. The overall average across the domains results in a minor improvement of +0.44.

We evaluate the best setup (the special domain markers) on the test set in order to confirm our findings. Following the trend on the development set, the improvement over the baseline is +2.19 Macro-F1.
The lower performance range of news over the other domains (both in dev and test) indicates that the different data source has a high impact even with the more uniform data distribution across domains proposed with our dataset extension.

\begin{table}[]
\centering
\resizebox{\columnwidth}{!}{
\begin{tabular}{c|lrrrrrrr}
\toprule
\multicolumn{2}{c}{} & \faNewspaper  & \faLandmark & \faLeaf & \faMusic & \faBookOpen & \faRobot & \textbf{avg.} \\
\midrule
\multirow{6}{*}{\rotatebox{90}{dev}}
 & \textsc{Baseline} & 25.45 & 31.35 & 39.46 & 39.69 & 38.84 & 38.09 & 35.48 \\
 \cmidrule{2-9}
 & \textsc{Dataset Emb.} & 15.38 & 22.22 & 24.77 & 32.64 & 30.95 & 29.80 & 25.96 \\
 & \textsc{Domain Mark.} & 26.36 & 32.77 & 40.31 & 42.65 & 40.59 & 38.71 & \textbf{36.90} \\
 & \textsc{Fine-grain.} & 23.67 & 32.67 & 35.35 & 38.76 & 38.23 & 35.94 & 34.10 \\
 & \textsc{Coarse-grain.} & 24.46 & 31.56 & 38.59 & 39.33 & 38.09 & 37.90 & 34.99 \\
 & \textsc{Dom. + Coarse} &  24.52 & 32.02 & 39.63 & 42.19 & 40.01 & 37.17 & 35.92 \\
 \midrule
 \multirow{2}{*}{\rotatebox{90}{test}}
 & \textsc{Baseline} & 24.73 & 34.12 & 39.67 & 39.96 & 44.64 & 35.71 & 36.47 \\
 \cmidrule{2-9}
 & \textsc{Domain Mark.} & 26.72 & 37.62 & 43.57 & 41.48 & 44.88 & 37.69 & \textbf{38.66} \\
\bottomrule
\end{tabular}}
\caption{\textbf{Performance Scores.} Macro-F1 scores of the explored setups. \textsc{Dom. + Coarse} indicates the combination of special domain markers with the coarse-grained entity types.}
\label{tab:results}
\end{table}

\begin{figure}
    \resizebox{0.99\columnwidth}{!}{
        \includegraphics{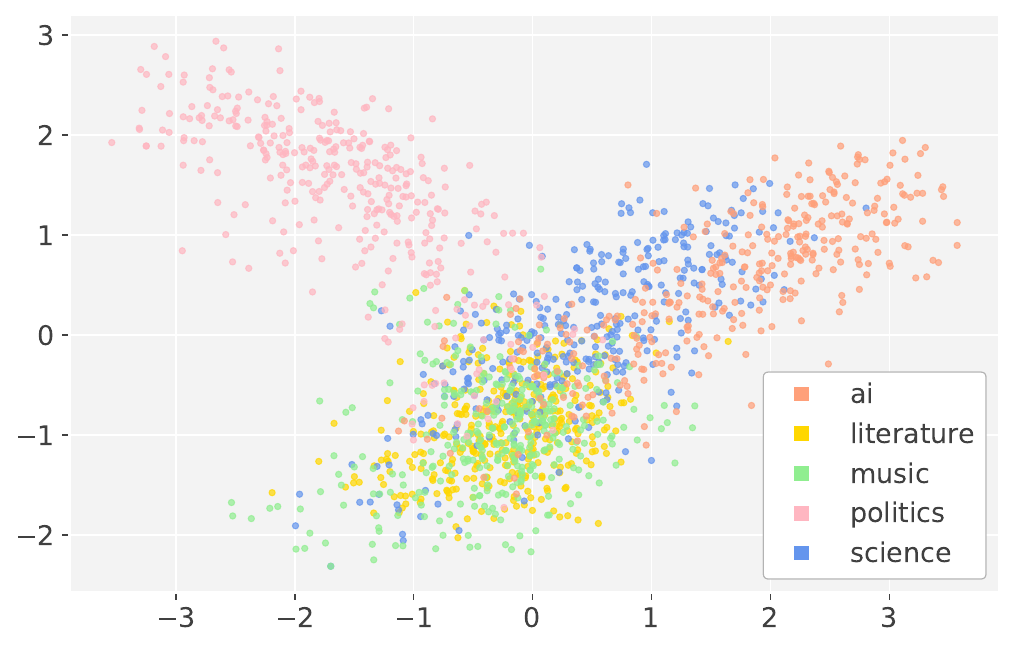}
    }
    \caption{\textbf{Domain Representation.} PCA plot of the untrained embeddings of the instances in the development set, colored by domain.}
    \label{fig:domains}
\end{figure}

\begin{figure*}
    \resizebox{0.99\textwidth}{!}{
        \includegraphics{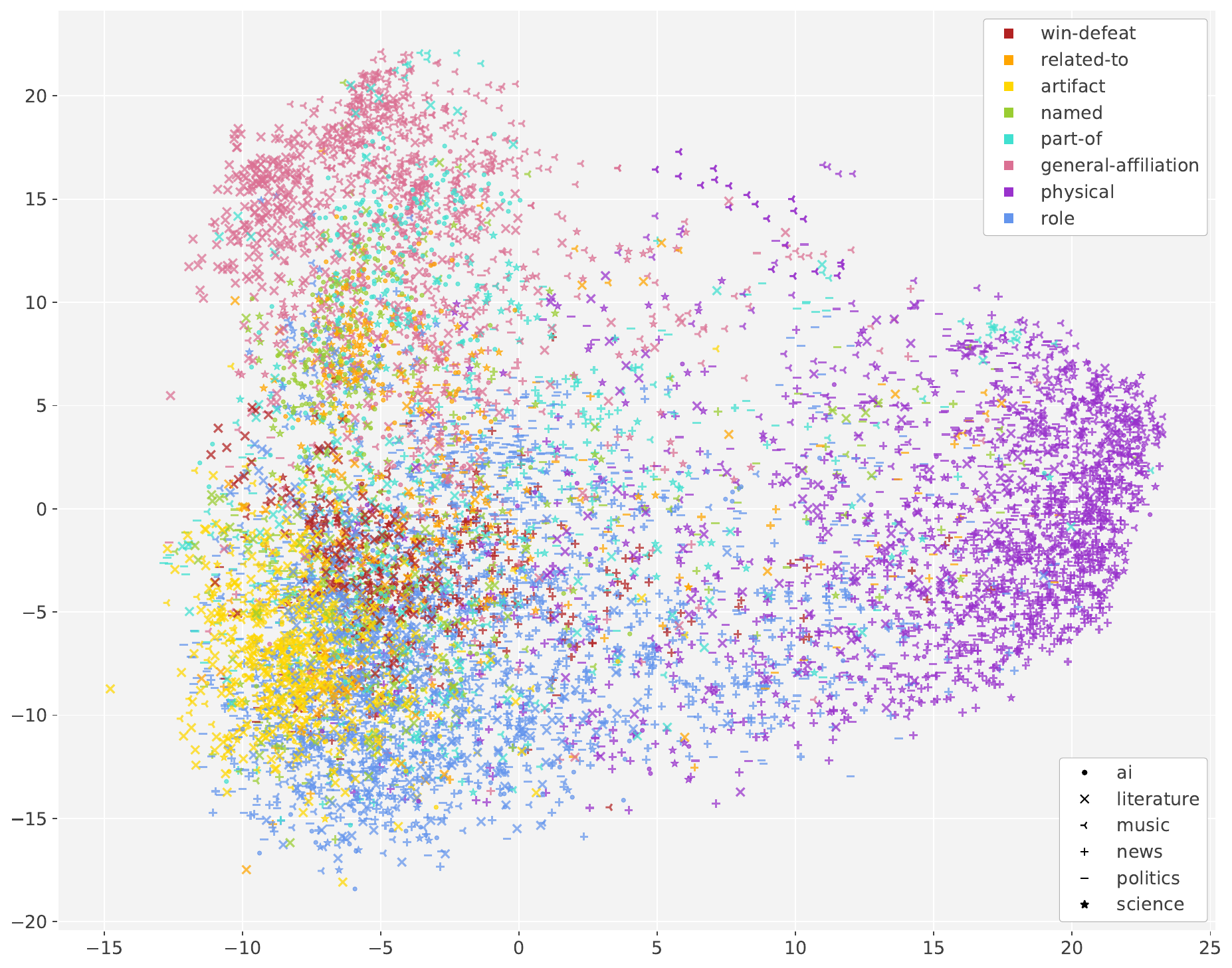}
    }
    \caption{\textbf{Relation Representation.} PCA plot of the trained embeddings of the most frequent relation labels in the development set, colored by relation labels and shaped by domain.}
    \label{fig:relations}
\end{figure*}

\section{Analysis}
\label{analysis}

\paragraph{Domain Representation} 
To inspect how much domain information the out-of-the-box embeddings contain, in Figure~\ref{fig:domains} we plot the PCA representations of the untrained embeddings (with \texttt{bert-base-cased}, the encoder used by the RC model) of the instances in the development set. 
We do not include the news domain in this plot because given its different data source (consisting of shorter sentences, typically news headlines), the news instances are very distant from the other domains, resulting in an high overlap of the latter. The current setup allows us to shed light on the remaining five domains, besides news which we already know is very distinctive.
The domains are already relatively distinguishable with the untrained encoder.
The two technical domains, science and AI, marginally overlap; politics is completely detached; and only music and literature overlap significantly.
Our intuition is that encoding additional domain information (see Section~\ref{sec:domains}) may not be particularly relevant.

\paragraph{Relation Representation}
We dive deeper into analyzing the relations and explore whether in the baseline setup (i.e., without additional domain information) the representation of instances coming from different domains, but belonging to the same class, are close to each other.
In Figure~\ref{fig:relations} we plot the PCA representation of the trained baseline model of the instances with the most frequent relation labels in the development set, separated by class and domain.
The main finding from this plot is that most of the classes are quite clustered, independently from the domain they belong to. For example, the \textit{physical} relation on the right side has instances from all the domains. Similarly, the \textit{artifact} and the \textit{role} labels towards the bottom-left corner of the plot.
Interestingly, the \textit{general-affiliation} relation presents clustered representations of the instances in the literature and music domains, but it still dominates the upper left side of the plot.
Less surprisingly, the \textit{related-to} label, listed as None-Of-The-Above (NOTA) in the guidelines, has a more sparse distribution across the plot.

The labels which present a less defined cluster (i.e., the ones whose meaning shifts the most across domains) are the ones which benefit the most from the special domain markers. For example, \textit{related-to} improves from a baseline value of 20.99 F1 up to 24.21 in the special domain markers setup; \textit{named} goes from 68.25 to 71.30 F1; and \textit{part-of} improves from 32.79 to 35.54 F1.
In contrary, the relation labels which present a better defined cluster already within the baseline (see Figure~\ref{fig:relations}) do not benefit much from the additional encoding of domain information. For example, the per-label F1 scores of the \textit{physical}, \textit{general-affiliation}, and \textit{role} relations in the baseline and special domain markers setups are respectively 77.16 and 77.51, 54.09 and 54.46, 65.60 and 65.11.

\section{Conclusion}

We explore how to encode domain information in a multi-domain training setup for the domain-specific task of RC. 
We propose CrossRE 2.0, a dataset extension of CrossRE~\cite{bassignana-plank-2022-crossre} for balancing the amount of data across the six domains included in it.
We manage to improve the multi-domain training baseline by $> 2$ Macro-F1 with a simple, but effective technique which encodes domain information in special domain markers concatenated at the beginning of each input.
Our analysis reveals that not all of the relation labels benefit the same from the domain encoding: The most generic, with a diverse interpretation across domains (e.g., \textit{part-of}) are the ones which gain the most in terms of per-label F1.

\section{Ethics Statement}

We do not foresee any potential risk related to this work. The data we use is published freely by~\citet{crossNER} and~\citet{bassignana-plank-2022-crossre}.

For the dataset extension, we hired an expert with a linguistics degree employed following national salary rates. The cost of the annotation process amounts to $\approx$ 1\$ per annotated relation.

\section*{Acknowledgments}
We thank our annotator for the
great job and substantial help given to this project. We thank the NLPnorth group at ITU and the MaiNLP group at LMU for feedback on an earlier version of this paper.
Elisa Bassignana and Barbara Plank are supported by the Independent Research Fund Denmark (Danmarks Frie Forskningsfond; DFF) Sapere Aude grant 9063-00077B. Barbara Plank is in parts supported by the European Research Council (ERC)  grant agreement No.\ 101043235.

\section{Bibliographical References}\label{sec:reference}

\bibliographystyle{lrec-coling2024-natbib}
\bibliography{anthology,custom}

\begin{thebibliography}{14}
\expandafter\ifx\csname natexlab\endcsname\relax\def\natexlab#1{#1}\fi

\bibitem[{Ammar et~al.(2016)Ammar, Mulcaire, Ballesteros, Dyer, and Smith}]{ammar-etal-2016-many}
Waleed Ammar, George Mulcaire, Miguel Ballesteros, Chris Dyer, and Noah~A. Smith. 2016.
\newblock \href {https://doi.org/10.1162/tacl_a_00109} {Many languages, one parser}.
\newblock \emph{Transactions of the Association for Computational Linguistics}, 4:431--444.

\bibitem[{Baldini~Soares et~al.(2019)Baldini~Soares, FitzGerald, Ling, and Kwiatkowski}]{baldini-soares-etal-2019-matching}
Livio Baldini~Soares, Nicholas FitzGerald, Jeffrey Ling, and Tom Kwiatkowski. 2019.
\newblock \href {https://doi.org/10.18653/v1/P19-1279} {Matching the blanks: Distributional similarity for relation learning}.
\newblock In \emph{Proceedings of the 57th Annual Meeting of the Association for Computational Linguistics}, pages 2895--2905, Florence, Italy. Association for Computational Linguistics.

\bibitem[{Bassignana and Plank(2022)}]{bassignana-plank-2022-crossre}
Elisa Bassignana and Barbara Plank. 2022.
\newblock \href {https://doi.org/10.18653/v1/2022.findings-emnlp.263} {{C}ross{RE}: A cross-domain dataset for relation extraction}.
\newblock In \emph{Findings of the Association for Computational Linguistics: EMNLP 2022}, pages 3592--3604, Abu Dhabi, United Arab Emirates. Association for Computational Linguistics.

\bibitem[{Conneau and Lample(2019)}]{xlmr}
Alexis Conneau and Guillaume Lample. 2019.
\newblock \href {https://proceedings.neurips.cc/paper_files/paper/2019/file/c04c19c2c2474dbf5f7ac4372c5b9af1-Paper.pdf} {Cross-lingual language model pretraining}.
\newblock In \emph{Advances in Neural Information Processing Systems}, volume~32. Curran Associates, Inc.

\bibitem[{Liu et~al.(2020)Liu, Gu, Goyal, Li, Edunov, Ghazvininejad, Lewis, and Zettlemoyer}]{liu-etal-2020-multilingual-denoising}
Yinhan Liu, Jiatao Gu, Naman Goyal, Xian Li, Sergey Edunov, Marjan Ghazvininejad, Mike Lewis, and Luke Zettlemoyer. 2020.
\newblock \href {https://doi.org/10.1162/tacl_a_00343} {Multilingual denoising pre-training for neural machine translation}.
\newblock \emph{Transactions of the Association for Computational Linguistics}, 8:726--742.

\bibitem[{Liu et~al.(2021)Liu, Xu, Yu, Dai, Ji, Cahyawijaya, Madotto, and Fung}]{crossNER}
Zihan Liu, Yan Xu, Tiezheng Yu, Wenliang Dai, Ziwei Ji, Samuel Cahyawijaya, Andrea Madotto, and Pascale Fung. 2021.
\newblock \href {https://ojs.aaai.org/index.php/AAAI/article/view/17587} {Crossner: Evaluating cross-domain named entity recognition}.
\newblock \emph{Proceedings of the AAAI Conference on Artificial Intelligence}, 35(15):13452--13460.

\bibitem[{Smith et~al.(2018)Smith, Bohnet, de~Lhoneux, Nivre, Shao, and Stymne}]{smith-etal-2018-82}
Aaron Smith, Bernd Bohnet, Miryam de~Lhoneux, Joakim Nivre, Yan Shao, and Sara Stymne. 2018.
\newblock \href {https://doi.org/10.18653/v1/K18-2011} {82 treebanks, 34 models: {U}niversal {D}ependency parsing with multi-treebank models}.
\newblock In \emph{Proceedings of the {C}o{NLL} 2018 Shared Task: Multilingual Parsing from Raw Text to Universal Dependencies}, pages 113--123, Brussels, Belgium. Association for Computational Linguistics.

\bibitem[{Stymne(2020)}]{stymne-2020-cross}
Sara Stymne. 2020.
\newblock \href {https://doi.org/10.18653/v1/2020.tlt-1.6} {Cross-lingual domain adaptation for dependency parsing}.
\newblock In \emph{Proceedings of the 19th International Workshop on Treebanks and Linguistic Theories}, pages 62--69, D{\"u}sseldorf, Germany. Association for Computational Linguistics.

\bibitem[{Stymne et~al.(2018)Stymne, de~Lhoneux, Smith, and Nivre}]{stymne-etal-2018-parser}
Sara Stymne, Miryam de~Lhoneux, Aaron Smith, and Joakim Nivre. 2018.
\newblock \href {https://doi.org/10.18653/v1/P18-2098} {Parser training with heterogeneous treebanks}.
\newblock In \emph{Proceedings of the 56th Annual Meeting of the Association for Computational Linguistics (Volume 2: Short Papers)}, pages 619--625, Melbourne, Australia. Association for Computational Linguistics.

\bibitem[{Tjong Kim~Sang and De~Meulder(2003)}]{tjong-kim-sang-de-meulder-2003-introduction}
Erik~F. Tjong Kim~Sang and Fien De~Meulder. 2003.
\newblock \href {https://aclanthology.org/W03-0419} {Introduction to the {C}o{NLL}-2003 shared task: Language-independent named entity recognition}.
\newblock In \emph{Proceedings of the Seventh Conference on Natural Language Learning at {HLT}-{NAACL} 2003}, pages 142--147.

\bibitem[{van~der Goot and de~Lhoneux(2021)}]{van-der-goot-de-lhoneux-2021-parsing}
Rob van~der Goot and Miryam de~Lhoneux. 2021.
\newblock \href {https://aclanthology.org/2021.tlt-1.9} {Parsing with pretrained language models, multiple datasets, and dataset embeddings}.
\newblock In \emph{Proceedings of the 20th International Workshop on Treebanks and Linguistic Theories (TLT, SyntaxFest 2021)}, pages 96--104, Sofia, Bulgaria. Association for Computational Linguistics.

\bibitem[{van~der Goot et~al.(2021)van~der Goot, {\"U}st{\"u}n, and Plank}]{van-der-goot-etal-2021-effectiveness}
Rob van~der Goot, Ahmet {\"U}st{\"u}n, and Barbara Plank. 2021.
\newblock \href {https://aclanthology.org/2021.adaptnlp-1.19} {On the effectiveness of dataset embeddings in mono-lingual,multi-lingual and zero-shot conditions}.
\newblock In \emph{Proceedings of the Second Workshop on Domain Adaptation for NLP}, pages 183--194, Kyiv, Ukraine. Association for Computational Linguistics.

\bibitem[{Wagner et~al.(2020)Wagner, Barry, and Foster}]{wagner-etal-2020-treebank}
Joachim Wagner, James Barry, and Jennifer Foster. 2020.
\newblock \href {https://doi.org/10.18653/v1/2020.acl-main.778} {Treebank embedding vectors for out-of-domain dependency parsing}.
\newblock In \emph{Proceedings of the 58th Annual Meeting of the Association for Computational Linguistics}, pages 8812--8818, Online. Association for Computational Linguistics.

\bibitem[{Zhong and Chen(2021)}]{zhong-chen-2021-frustratingly}
Zexuan Zhong and Danqi Chen. 2021.
\newblock \href {https://doi.org/10.18653/v1/2021.naacl-main.5} {A frustratingly easy approach for entity and relation extraction}.
\newblock In \emph{Proceedings of the 2021 Conference of the North American Chapter of the Association for Computational Linguistics: Human Language Technologies}, pages 50--61, Online. Association for Computational Linguistics.

\end{thebibliography}


\end{document}